\documentclass[11pt,a4paper]{article}
\usepackage[hyperref]{acl2017}
\usepackage{times}
\usepackage{url}
\usepackage{latexsym}

\usepackage{graphicx}
\usepackage{multirow}
\usepackage{amsmath}

\aclfinalcopy

\title{Improving Correlation with Human Judgments\\
by Integrating Semantic Similarity with Second--Order Vectors} 

\author{Bridget T. McInnes\\
	    Department of Computer Science\\
	    Virginia Commonwealth University\\
	    Richmond, VA 23284 USA\\
	    {\tt btmcinnes@vcu.edu}
	  \And
	Ted Pedersen\\
  	Department of Computer Science\\
  	University of Minnesota\\
  	Duluth, MN 55812 USA\\
	  {\tt tpederse@d.umn.edu}}

\begin{document}

\maketitle

\begin{abstract}
Vector space methods that measure semantic similarity and relatedness 
often rely on distributional information such as co--occurrence frequencies or statistical measures of association to weight the importance of
particular co--occurrences. In this paper, we extend these methods by incorporating a measure of semantic similarity based on a human curated taxonomy into a second--order vector representation. This results in a measure of semantic relatedness that combines both the contextual  information available in a corpus--based vector space representation with the semantic knowledge found in a biomedical ontology. Our results show that incorporating semantic similarity into a second order co--occurrence matrices improves correlation with human judgments for both similarity and relatedness, and that our method compares favorably to various different word embedding methods that have recently been evaluated on the same reference standards we have used.
\end{abstract}

\section{Introduction}

Measures of semantic similarity and relatedness quantify the degree to which 
two concepts are similar (e.g., $lung$--$heart$) or related (e.g., $lung$--
$bronchitis$). Semantic similarity can be viewed as a special case of semantic relatedness -- to be similar is one of many ways that a pair of concepts may be related. The automated discovery of groups of semantically similar or  related terms is critical to improving the retrieval~\cite{RadaMBB89} and clustering~\cite{LinLCL07} of biomedical and clinical documents, and the development of biomedical terminologies and ontologies~\cite{BodenreiderB04}. 

There is a long history in using distributional methods to discover semantic similarity and relatedness (e.g., \cite{LinP02,ReisingerM10,RadinskyAGM11,YihQ12}). These methods are all based on the distributional hypothesis, which holds that two terms that are distributionally similar (i.e., used in the same context) will also be semantically similar \cite{Harris54,WeedsWM04}. Recently word embedding techniques such as word2vec \cite{mikolov2013distributed} have become very popular. Despite the prominent role that neural networks play in many of these approaches, at their core they remain distributional techniques that typically start with a word by word co--occurrence matrix, much  like many of the more traditional approaches. 

However, despite these successes distributional methods do not perform well when data is very sparse (which is common). One possible solution is to use second--order co--occurrence vectors \cite{Schutze92,Schutze98}. In this approach the  similarity between two words is not  strictly based on their co--occurrence frequencies, but rather on the frequencies of the other words which occur with both of them (i.e., second order co--occurrences). This approach has been shown to be successful in quantifying semantic relatedness~\cite{IslamI06,PedersenPPC07}. However, while more robust in
the face of sparsity, second--order methods can result in significant amounts of noise, where contextual information that is overly general is included and does not contribute to quantifying the semantic relatedness between the two concepts.

Our goal then is to discover methods that automatically reduce the amount of
noise in a second--order co--occurrence vector. We achieve this by incorporating pairwise semantic similarity scores derived from a taxonomy into our second--order vectors, and then using these scores to select only the most semantically similar co--occurrences (thereby reducing noise).

We evaluate our method on two datasets that have been annotated in multiple ways.  One has been annotated for both similarity and relatedness, and the other has been annotated for relatedness by two different types of experts (medical doctors and medical coders). Our results show that integrating second order co--occurrences with measures of semantic similarity increases correlation with our human reference standards. We also compare our result to a number of other studies which have applied various word embedding methods to the same reference standards we have used. We find that our method often performs at a comparable or higher level than these approaches. These results suggest that our methods of integrating semantic similarity and relatedness values have the  potential to improve performance of purely distributional methods. 

\section{Similarity and Relatedness Measures}
\label{sec:measures}

This section describes the similarity and relatedness measures we integrate in our second--order co--occurrence vectors. We use two taxonomies in this study, SNOMED--CT and MeSH. SNOMED--CT ({\it Systematized Nomenclature of Medicine Clinical Terms}) is a comprehensive clinical terminology created for the electronic representation of clinical health information. MeSH ({\it Medical Subject Headings}) is a taxonomy of biomedical terms developed for indexing biomedical journal articles. 

We obtain SNOMED--CT and MeSH via the Unified Medical Language System (UMLS) Metathesaurus (version 2016AA). The Metathesaurus contains approximately 2 million biomedical and clinical concepts from over 150 different terminologies that have been semi--automatically integrated into a single source. Concepts in the Metathesaurus are connected largely
by two types of hierarchical relations: $parent$/$child$ (PAR/CHD) and $broader$/$narrower$ (RB/RN). 

\subsection{Similarity Measures}
\label{sec:similarity}

Measures of semantic similarity can be classified into three broad categories : path--based, feature--based and information content (IC). Path--based similarity measures use the structure of a taxonomy to measure similarity -- concepts positioned close to each other are more similar than
those further apart. Feature--based methods rely on set theoretic measures
of overlap between features (union and intersection). The information
content measures quantify the amount of information that a concept
provides -- more specific concepts have a higher amount of information
content. 

\subsubsection{Path--based Measures}

\newcite{RadaMBB89} introduce the {\em Conceptual Distance} measure. This 
measure is simply the length of the shortest path between two concepts 
($c1$ and $c2$) in the MeSH hierarchy. Paths are based on {\em broader than}
 (RB)
and {\em narrower than} (RN) relations. \newcite{CaviedesC04} extends this
measure to use {\em parent} (PAR) and {\em child} (CHD) relations. 
Our $path$ measure is simply the reciprocal of this shortest path 
value (Equation \ref{eq:path}), so that larger values 
(approaching 1) indicate a high degree of similarity. 

\begin{equation}
path = \frac{1}{spath(c_1,c_2)}
\label{eq:path}
\end{equation}

While the simplicity of $path$ is appealing, it can be misleading
when concepts are at different levels of specificity. Two very general
concepts may have the same path length as two very specific concepts.
\newcite{WuP94} introduce a correction to $path$ that incorporates the
depth of the concepts, and the depth of their Least Common Subsumer (LCS).
This is the most specific ancestor two concepts share.
In this measure, similarity is twice the depth of the two concept's 
LCS divided by the product of the depths of the individual concepts 
(Equation \ref{eq:wup}). Note that if there are multiple LCSs for a pair 
of concepts, the deepest of them is used in this measure. 

\begin{equation}
wup = \frac {2 * depth(lcs(c_1, c_2))} {depth(c_1) + depth(c_2)}
\label{eq:wup}
\end{equation}

\newcite{ZhongZLY02} take a very similar approach and again scale the
depth of the LCS by the sum of the depths of the two concepts 
(Equation~\ref{eq:zhong}), 
where $m(c) = k^{-depth(c)}$. The value of $k$ was set to 2 based on
their recommendations. 

\begin{equation}
zhong = \frac{2 * m(lcs(c_1,c_2))}{m(c_1) + m(c_2)} 
\label{eq:zhong}
\end{equation}

\newcite{PekarS02} offer another variation on $path$, where the shortest
path of the two concepts to the LCS is used, in addition to the 
shortest bath between the LCS and the root of the taxonomy 
(Equation~\ref{eq:pks}).

\begin{equation}
\begin{aligned}
pks = -\log \frac{spath(lcs(c_1,c_2), root)}{\sum_{x=c_1, c_2, root} spath(lcs(c_1,c_2), x)}
\label{eq:pks}
\end{aligned}
\end{equation}

\subsubsection{Feature--based Measures}

Feature--based methods represent each concept as a set of features and then 
measure the overlap or sharing of features to measure
similarity. In particular, each concept is represented as the set of their
ancestors, and similarity is a ratio of the intersection and union of these 
features. 

\newcite{MaedcheS01} quantify the similarity between two concepts 
as the ratio of the intersection over their union as shown in Equation~\ref{eq:cmatch}. 

\begin{equation}
cmatch = \frac{|A(c_1)\bigcap A(c_2)|}{|A(c_1)\bigcup A(c_2)|}
\label{eq:cmatch}
\end{equation}

\newcite{BatetSV11} extend this by excluding any shared features (in the numerator) as shown in Equation~\ref{eq:batet}. 

\begin{equation}
batet = -log_2(\frac{|A(c_{1})\bigcup A(c_2)| - |A(c_1)\bigcap A(c_2)|}{|A(c_1) \bigcup A(c_2)|})
\label{eq:batet}
\end{equation}

\subsubsection{Information Content Measures}
 
Information content is formally defined as the negative log of the probability of a concept.
The effect of this is to assign rare (low probability) concepts a high measure of information
content, since the underlying assumption is that more specific concepts are
less frequently used than more common ones. 

\newcite{Resnik95} modified this notion of information content in order to use it as a similarity
measure. He defines the similarity of two concepts to be the information content of their LCS
(Equation~\ref{eq:res}).  

\begin{equation}
res = IC(lcs(c_1, c_2) = -\log(P(lcs(c_1, c_2)))
\label{eq:res}
\end{equation}

\newcite{JiangC97}, \newcite{Lin98}, and \newcite{PirroE10}  
extend $res$ by incorporating the information content of the individual 
concepts in various different ways. \newcite{Lin98} defines the similarity between two concepts
as the ratio of information content of the LCS with the sum of the individual concept's information 
content (Equation~\ref{eq:lin}). Note that $lin$ has the same form as $wup$ and $zhong$, 
and is in effect using
information content as a measure of specificity (rather than depth). 
If there is more than one possible LCS, the LCS with the greatest IC is chosen. 

\begin{equation}
lin = \frac{2*IC(lcs(c_{1},c_{2}))}{IC(c_{1}) + IC(c_{2})}
\label{eq:lin}
\end{equation}

\newcite{JiangC97} define the distance between two concepts to be the 
sum of the information content
 of the two concepts minus twice the information content
 of the concepts' LCS. 
We modify this from a distance to a similarity measure by taking the 
reciprocal of the distance (Equation~\ref{eq:jcn}). Note that the denominator
of $jcn$ is very similar to the numerator of $batet$. 

\begin{equation}
jcn = \frac{1}{IC(c_1) + IC(c_2) - 2*IC(lcs(c_1,c_2))}
\label{eq:jcn}
\end{equation}

\newcite{PirroE10} 
define the similarity between two concepts as the 
information content of the two concept's LCS divided by the sum of their individual information
content values 
minus the information content of their LCS (Equation~\ref{eq:faith}). Note that $batet$ can be
viewed as a set--theoretic version of $faith$. 

\begin{equation}
faith = \frac{IC(lcs(c_1,c_2))}{IC(c_1) + IC(c_2) - IC(lcs(c_1,c_2))}
\label{eq:faith}
\end{equation}

\subsection{Information Content}

\label{sec:IC}
The information content of a concept may be 
derived from a corpus (corpus--based) or directly from a 
taxonomy (intrinsic--based). In this work we focus on 
corpus--based techniques.

For corpus--based information content, 
we estimate the probability of a concept $c$ by taking the sum of 
the probability of the concept $P(c)$ and the 
probability its descendants $P(d)$ (Equation~\ref{eq:prop}).

\begin{equation}
  P(c*) = P(c) + \sum_{d\in descendant(c)} P(d)
  \label{eq:prop}
\end{equation}

The initial probabilities of a concept ($P(c)$) and its descendants 
($P(d)$) are obtained by dividing the number of times each concept
and descendant occurs in the corpus, and dividing that by the total 
numbers of concepts ($N$). 

Ideally the corpus from which we are estimating the probabilities of 
concepts will be sense--tagged. However, sense--tagging is a challenging
problem in its own right, and it is not always possible to carry out reliably
on larger amounts of text. In fact in this paper we did not use any 
sense--tagging of the corpus we derived information content from. 

Instead, we estimated the probability of a concept by using the 
{\it UMLSonMedline} dataset. This was created by the National Library
of Medicine and consists of concepts from the 2009AB UMLS and the counts of
the number 
of times they occurred in a snapshot of Medline taken on 12 January, 2009. 
These counts were obtained by using the Essie Search Engine 
\cite{IdeLD07} which queried Medline with normalized strings 
from the 2009AB MRCONSO table in the UMLS. The frequency of a CUI was obtained 
by aggregating the frequency counts of the terms associated with the CUI to 
provide a rough estimate of its frequency. The information content measures
then use this information to calculate the probability of a concept. 

Another alternative is the use of {\em Intrinsic Information Content}. 
It assess the informativeness of concept based on its placement 
within a taxonomy by considering the number of incoming (ancestors) relative to
outgoing (descendant) links \cite{SanchezBI11} (Equation~\ref{eq:sanchez}).

\begin{equation}
  IC(c) = -log(\frac{\frac{|leaves(c)|}{|subsumers(c)|} + 1}{max\_leaves + 1})
  \label{eq:sanchez}
\end{equation}

\noindent 
where $leaves$ are the number of descendants of concept $c$ that
are leaf nodes, $subsumers$ are the number of concept $c$'s ancestors
and $max\_leaves$ are the total number of leaf nodes in the taxonomy.

\subsection{Relatedness Measures}
\label{sec:relatedness}
%lesk

\newcite{Lesk86} observed that concepts that are related should share more 
words in their respective definitions than concepts that are less connected. 
He was able to perform
word sense disambiguation by identifying the senses of words in a sentence
with the largest number of overlaps between their definitions.  
An overlap is the longest sequence of one 
or more consecutive words that occur in both definitions. 
\newcite{BanerjeeP03} extended this idea to WordNet, but observed that
WordNet glosses are often very short, and did not contain enough information to
distinguish between multiple concepts.  Therefore, they created a
{\em super--gloss} for each concept by adding the glosses of related
concepts to the gloss of the concept itself (and then finding overlaps).

\newcite{PatwardhanP06} adapted this measure to second--order co--occurrence 
vectors. In this approach, a vector is created for 
each word in a concept's definition that shows which words co--occur
with it in a corpus. These word vectors are averaged to create a 
single co-occurrence vector for the concept. The similarity between 
the concepts is calculated by taking the cosine between the concepts 
second--order vectors. \newcite{LiuMPMP12} modified and extended 
this measure to be used to quantify the relatedness between biomedical 
and clinical terms in the UMLS. The work in this paper can be seen
as a further extension of \newcite{PatwardhanP06} and \newcite{LiuMPMP12}. 

\section{Method}
\label{sec:method}

In this section, we describe our {\it second--order similarity vector} 
measure. This incorporates both contextual information using the term 
pair's definition and their pairwise semantic similarity scores derived 
from a taxonomy. There are two stages to our approach. First, a co--occurrence matrix must be constructed. Second, this matrix is used to construct a second--order co--occurrence vector for each concept in a pair of concepts to be measured for relatedness. 

\subsection{Co--occurrence Matrix Construction}

We build an $m{\times}n$ similarity matrix using an external corpus where the rows and columns represent words within the corpus and the element contains the similarity score between the row word and column word using the similarity measures discussed above. If a word maps to more than one possible sense, we use the sense that returns the highest similarity score. 

For this paper our external corpus was the NLM 2015 Medline baseline. Medline is a bibliographic database containing over 23 million citations to journal articles in the biomedical domain and is maintained by National Library of Medicine. The 2015 Medline Baseline encompasses approximately 5,600 journals starting from 1948 and contains 23,343,329 citations, of which 2,579,239 contain abstracts. In this work, we use Medline titles and abstracts from 1975 to present day. Prior to 1975, only 2\% of the citations contained an abstract. We then calculate the similarity for each bigram in this dataset and include those that have a similarity score greater than a specified threshold on these experiments. 

\subsection{Measure Term Pairs for Relatedness}

We obtain definitions for each of the two terms we wish to measure. Due to the sparsity and inconsistencies of the definitions in the UMLS, we not only use the definition of the term (CUI) but also include the definition of its related concepts. This follows the method proposed by \newcite{PatwardhanP06} for general English and WordNet, and which was adapted for the UMLS and the medical domain by \newcite{LiuMPMP12}. In particular we add the definitions of any concepts connected via a parent (PAR), child (CHD), RB (broader than), RN (narrower than) or TERM (terms associated with CUI) relation. All of the definitions for a term are combined into a single {\em super--gloss}. At the end of this process we should have two super--glosses, one for each term to be measured for relatedness. 

Next, we process each super--gloss as follows: 

\begin{enumerate}

\item We extract a first--order co--occurrence vector for each term in the super--gloss from the co--occurrence matrix created previously. 

\item We take the average of the first order co--occurrence vectors associated with the terms in a super--gloss and use that to represent the meaning of the term. This is a second--order co--occurrence vector. 

\item After a second--order co--occurrence vector has been constructed for each term, then we calculate the cosine between these two vectors to measure the relatedness of the terms. 

\end{enumerate}

\section{Data}

We use two reference 
standards
to evaluate the semantic similarity and relatedness measures 
\footnote{http://www.people.vcu.edu/~btmcinnes/downloads.html}.
UMNSRS was annotated for both similarity and relatedness by medical 
residents. MiniMayoSRS was annotated
for relatedness by medical doctors (MD) and medical coders (coder). 
In this section, we describe these data sets and describe a few of their differences.

{\bf MiniMayoSRS}: The MayoSRS, developed by \newcite{PakhomovPMMRC10}, consists of 101 clinical term pairs whose relatedness was determined by nine medical coders and three physicians from the Mayo Clinic. The relatedness of each term pair was assessed based on a four point scale: (4.0) practically synonymous, (3.0) related, (2.0) marginally related and (1.0) unrelated. MiniMayoSRS is a subset of the MayoSRS and consists of 30 term pairs on which a higher inter--annotator agreement was achieved. The average correlation between physicians is 0.68. The average correlation between medical coders is 0.78. 
We evaluate our method on the mean of the physician scores, and the mean 
of the coders scores in this subset in the same manner as reported by \newcite{PedersenPPC07}. 

{\bf UMNSRS}:  The University of Minnesota Semantic Relatedness Set (UMNSRS) 
was
developed by \newcite{PakhomovMALPM10}, and consists 
of 725 clinical term pairs whose semantic similarity and relatedness was 
determined independently by four medical residents from the University 
of Minnesota Medical School. The similarity and relatedness of each 
term pair was annotated based on a continuous scale by having the 
resident touch a bar on a touch sensitive computer screen to indicate 
the degree of similarity or relatedness. The Intraclass Correlation 
Coefficient (ICC) for the reference standard tagged for similarity 
was 0.47, and 0.50 for relatedness. Therefore, as suggested by Pakhomov 
and colleagues,we use a subset of the ratings consisting of 401 pairs 
for the similarity set and 430 pairs for the relatedness set which each 
have an ICC of 0.73. 

\section{Experimental Framework}
\label{sec:experimentalframework}

We conducted our experiments using the freely available open source 
software package UMLS::Similarity~\cite{McInnesPP09} 
version 1.47\footnote{http://search.cpan.org/edist/UMLS-Similarity/}.
This package takes as input two terms (or UMLS concepts) and returns 
their similarity or relatedness using the measures discussed in 
Section~\ref{sec:measures}. 

Correlation between the similarity measures and human 
judgments were estimated using Spearman's Rank Correlation ($\rho$). 
Spearman's measures the statistical dependence between two variables to 
assess how well the relationship between the rankings of the variables 
can be described using a monotonic function. 
We used Fisher's r-to-z transformation~\cite{Fisher15} to calculate the 
significance between the correlation results. 

\section{Results and Discussion}
\label{sec:results} 

Table~\ref{tbl:correlationresults} shows the Spearman's Rank Correlation between the human scores from the four reference standards and the scores from the various measures of similarity introduced in Section \ref{sec:measures}. Each class of measure is followed by the scores obtained when integrating our second order vector approach with these measures of semantic similarity.

\begin{table}
\begin{center}
\caption{Spearman's Correlation Results}
\begin{tabular}{|l|cc|cc|}
\hline
 & \multicolumn{2}{|c|}{UMNSRS} & \multicolumn{2}{|c|}{MiniMayoSRS} \\
 & \multicolumn{2}{|c|}{Resident} & \multicolumn{1}{|c}{MD} & \multicolumn{1}{c|}{Coder} \\
 & \multicolumn{1}{|c}{sim} & \multicolumn{1}{c|}{rel} & \multicolumn{2}{|c|}{relatedness} \\
\hline
\multicolumn{1}{|c|}{Path} & \multicolumn{2}{|c|}{} & \multicolumn{2}{|c|}{} \\
 path    & 0.52   & 0.28   & 0.35   & 0.45     \\
 wup     & 0.50   & 0.24   & 0.39   & 0.51     \\
 pks     & 0.49   & 0.25   & 0.38   & 0.50     \\
 zhong   & 0.50   & 0.25   & 0.42   & 0.50     \\
\multicolumn{1}{|c|}{{\it Integrated}} & \multicolumn{2}{|c|}{} & \multicolumn{2}{|c|}{} \\
 vector-path     & 0.60       & 0.43   & 0.54   & 0.54         \\
 vector-wup      & 0.60       & 0.42   & 0.55   & 0.55         \\
 vector-pks      & 0.60       & 0.42   & 0.53   & 0.53         \\
 vector-zhong    & 0.58       & 0.41   & 0.54   & 0.53         \\
\hline
\multicolumn{1}{|c|}{Feature} & \multicolumn{2}{|c|}{} & \multicolumn{2}{|c|}{} \\
  batet   & 0.16   & 0.33   & 0.16   & 0.15     \\  
 cmatch  & 0.33   & 0.17   & 0.35   & 0.35     \\
\multicolumn{1}{|c|}{{\it Integrated}} & \multicolumn{2}{|c|}{} & \multicolumn{2}{|c|}{} \\
 vector-batet    & 0.59       & 0.43   & 0.53   & 0.51        \\
 vector-cmatch   & 0.60       & 0.43   & 0.54   & 0.55         \\
\hline
\multicolumn{1}{|c|}{IC} & \multicolumn{2}{|c|}{} & \multicolumn{2}{|c|}{} \\
 res     & 0.49   & 0.26   & 0.36   & 0.47     \\
 lin     & 0.51   & 0.29   & 0.44   & 0.54     \\
 jcn     & 0.52   & 0.33   & 0.42   & 0.52     \\
 faith   & 0.51   & 0.29   & 0.43   & 0.54     \\
\multicolumn{1}{|c|}{{\it Integrated}} & \multicolumn{2}{|c|}{} & \multicolumn{2}{|c|}{} \\
 {\bf vector-res}& 0.57       & 0.41   & 0.58   & {\bf 0.65} \\
 vector-lin      & 0.57       & 0.41   & 0.59   & 0.64        \\
 vector-jcn      & 0.42       & 0.15   & 0.26   & 0.41        \\
 {\bf vector-faith}& {\bf 0.59} & 0.42   & 0.58   & 0.63      \\
\hline
\multicolumn{1}{|c|}{Intrinsic IC} & \multicolumn{2}{|c|}{} & \multicolumn{2}{|c|}{} \\
 ires    & 0.49   & 0.26   & 0.40   & 0.50  \\
 ilin    & 0.50   & 0.28   & 0.41   & 0.50   \\
 ijcn    & 0.51   & 0.29   & 0.39   & 0.50   \\
 ifaith  & 0.50   & 0.28   & 0.41   & 0.50   \\
\multicolumn{1}{|c|}{{\it Integrated}} & \multicolumn{2}{|c|}{} & \multicolumn{2}{|c|}{} \\
 vector-ires     & 0.57       & 0.41   & 0.50   & 0.52      \\
 vector-ilin     & 0.57       & 0.41   & 0.55   & 0.59      \\
 vector-ijcn     & 0.50       & 0.41   & 0.54   & 0.54       \\
 vector-ifaith   & 0.58       & 0.42   & 0.58   & 0.64       \\
\hline
\multicolumn{1}{|c|}{Relatedness} & \multicolumn{2}{|c|}{} & \multicolumn{2}{|c|}{} \\
 lesk     & 0.49   & 0.33   & 0.52   & 0.56    \\
 o1vector & 0.47   & 0.36   & 0.43   & 0.54 \\
 $o2vector$ & 0.54   & {\bf 0.45} & {\bf 0.63} & 0.59   \\
\hline
\end{tabular}
\label{tbl:correlationresults}
\end{center}
\end{table} 

\subsection{Results Comparison}

The results for UMNSRS tagged for similarity ($sim$) and 
MiniMayoSRS tagged by coders show that all of the second-order 
similarity vector measures ($Integrated$) except for $vector$-$jcn$ obtain 
a higher correlation than the original measures. We found that  
$vector$-$res$ and $vector$-$faith$ obtain the highest correlations
of all these results with human judgments.

For the UMNSRS dataset tagged for relatedness and MiniMayoSRS 
tagged by physicians (MD), the original $vector$ measure obtains a
higher correlation than our measure ($Integrated$) although the difference
is not statistically significant ($p \le 0.2$). 

In order to analyze and better understand these results, we filtered 
the bigram pairs used to create 
the initial similarity matrix based on the strength of their similarity using 
the $faith$ and the $res$ measures. Note that the $faith$ measure holds to
a 0 to 1 scale, while $res$ ranges from 0 to an unspecified upper bound that
is dependent on the size of the corpus from which information content is 
estimated. As such we use a different range of threshold values for each
measure. We discuss the results of this filtering below. 

\subsection {Thresholding Experiments}

Table~\ref{tbl:res} shows the results of applying the threshold 
parameter on each of the reference standards using the $res$ 
measure. For example, a threshold of $0$ indicates that all 
of the bigrams were included in the similarity matrix; and a threshold 
of $1$ indicates that only the bigram pairs with a similarity score 
greater than one were included. 

These results show that using a threshold cutoff of $2$ obtains the 
highest correlation for the UMNSRS dataset, and that a 
threshold cutoff of $4$ obtains the highest correlation for the 
MiniMayoSRS dataset. All of the results show an increase in correlation 
with human judgments when incorporating a threshold cutoff over all 
of the original measures.  
The increase in the correlation for the UMNSRS tagged for similarity  
 is statistically significant ($p \le 0.05$), however this is not 
the case for the UMNSRS tagged for relatedness nor for the MiniMayoSRS 
data. 

\begin{table}
\begin{center}
\caption{Threshold Correlation with $vector$-$res$} 
\begin{tabular}{|c|r|cc|cc|}
\hline
  &      & \multicolumn{2}{|c|}{UMNSRS} & \multicolumn{2}{|c|}{MiniMayoSRS} \\
T & \# bigrams  & sim & rel & MD & coder \\
\hline
0 & 850,959 & 0.58       & 0.41       & 0.58       & 0.65    \\
1 & 166,003 & 0.56       & 0.39       & 0.60       & 0.67     \\
2 & 65,502  & {\bf 0.64} & {\bf 0.47} & 0.56       & 0.62     \\
3 & 27,744  & 0.60       & 0.46       & 0.62       & 0.71     \\
4 & 10,991  & 0.56       & 0.43       & {\bf 0.75} & {\bf 0.76} \\
5 & 3,305   & 0.26       & 0.16       & 0.36       & 0.36       \\
\hline
\end{tabular}
\label{tbl:res}
\end{center}
\end{table} 

Similarly, Table~\ref{tbl:faith} shows the results of applying the threshold 
parameter (T) on each of the reference standards using the $faith$ measure. Although, 
unlike $res$ whose scores are greater than or equal to 0 without an upper limit, the $faith$ measure 
returns scores between 0 and 1 (inclusive). Therefore, here a threshold of $0$ indicates 
that all of the bigrams were included in the similarity matrix; and a threshold 
of $0.1$ indicates that only the bigram pairs with a similarity score 
greater than $0.1$ were included. The results show an increase in accuracy 
for all of the datasets except for the MiniMayoSRS tagged for physicians. 
The increase in the results for the UMNSRS tagged for similarity and the 
MayoSRS is statistically significant ($p \le 0.05$). This is not 
the case for the UMNSRS tagged for relatedness nor the MiniMayoSRS.

\begin{table}
\begin{center}
\caption{Threshold Correlation with $vector$-$faith$} 
\begin{tabular}{|c|r|cc|cc|}
\hline
  & \#       & \multicolumn{2}{|c|}{UMNSRS} & \multicolumn{2}{|c|}{MiniMayoSRS} \\
T & bigrams  & sim & rel & MD & coder \\
\hline
%%0   & 838,353 & 0.54       & 0.45       & {\bf 0.63}& 0.59         \\
0   & 838,353 & 0.59       & 0.42       & {\bf 0.58} & {\bf 0.63}         \\
0.1 & 197,189 & 0.58       & 0.41       & 0.57       & {\bf 0.63}        \\
0.2 & 121,839 & 0.58       & 0.41       & {\bf 0.58}       & {\bf 0.63}  \\
0.3 & 71,353  & 0.63       & 0.46       & 0.54       & 0.55        \\
0.4 & 45,335  & 0.64       & 0.48       & 0.50       & 0.51        \\
0.5 & 29,734  & {\bf 0.66} & {\bf 0.49} & 0.49       & 0.53        \\
0.6 & 19,347  & 0.65       & 0.49       & 0.52       & 0.56        \\
0.7 & 11,946  & 0.64       & 0.48       & 0.53       & 0.55        \\
0.8 & 7,349   & 0.64       & 0.49       & 0.53       & 0.56        \\
0.9 & 4,731   & 0.62       & 0.49       & 0.53       & 0.57        \\
\hline
\end{tabular}
\label{tbl:faith}
\end{center}
\end{table} 

Overall, these results indicate that including only those bigrams that have 
a sufficiently high similarity score increases the correlation 
results with human judgments, but what quantifies as sufficiently 
high varies depending on the dataset and measure.  

\begin{table*}
\begin{center}
\caption{Comparison with Previous Work} 
\scalebox{0.8}{
\begin{tabular}{|l|cc|cc|c|ccc|}
\hline
 Method & \multicolumn{4}{|c|}{UMNSRS} & MayoSRS  & \multicolumn{3}{|c|}{MiniMayoSRS} \\
        & \multicolumn{2}{|c|}{Subsets} & \multicolumn{2}{|c|}{Full} & (N=101) & \multicolumn{3}{|c|}{(N=29)} \\
  & sim & rel & sim (N=566) & rel (N=587) & rel & MD & coder & avg\\
\hline
vector--res (ours)           & 0.64 (N=401) & 0.49 (N=430)   &  0.59 & 0.48  & 0.51 & 0.75  & 0.76 & 0.76 \\
vector--faith (ours)         & 0.66 (N=401) & 0.49 (N=430)  &  0.61 & 0.49  & 0.46 & 0.58  & 0.63 & 0.63 \\
\hline
\cite{YuCBJW16}              &       &       &       &       &       & 0.70 & 0.67 & \\
\cite{sajad2015domain}       &        &      & 0.39  & 0.39  &  0.63 &       &       & 0.8 \\
\hline
\cite{Pakhomov2016corpus}    & 0.62 (N=449) & 0.58 (N=458) & & & & & & \\
\cite{muneeb2015evalutating} & 0.52 (N=462) & 0.45 (N=465) & & & & & & \\
\cite{chiu2016how}           & 0.65 (N=UK)  & 0.60 (N=UK) & & & & & &\\
\hline
\end{tabular}
}
\label{tbl:relatedwork}
\end{center}
\end{table*} 

\subsection{Comparison with Previous Work}

Recently, word embeddings~\cite{mikolov2013distributed} have become a popular method for measuring semantic relatedness in the biomedical domain. This is a neural network based approach that learns a representation of a word by word co--occurrence matrix. The basic idea is that the neural network learns a series of weights (the hidden layer within the neural network) that either maximizes the probability of a word given its context, referred to as the continuous bag of words (CBOW) approach, or that maximizes the probability of the context given a word, referred to as the Skip--gram approach.  These approaches have been used in numerous recent papers. 

\newcite{muneeb2015evalutating} trained both the Skip--gram and CBOW models over the PubMed Central Open Access (PMC) corpus of approximately 1.25 million articles. They evaluated the models on a subset of the UMNSRS data, removing word pairs that did not occur in their training corpus more than ten times. \newcite{chiu2016how} evaluated both the  the Skip--gram and CBOW models over the PMC corpus and PubMed. They also evaluated the models on a subset of the UMNSRS ignoring those words that did not appear in their training corpus.  \newcite{Pakhomov2016corpus} trained CBOW model over three different types of corpora: clinical (clinical notes from the Fairview Health System), biomedical (PMC corpus), and general English (Wikipedia). They evaluated their method using a subset of the UMNSRS restricting to single word term pairs and removing those not found within their training corpus. \newcite{sajad2015domain} trained the Skip--gram model over CUIs identified by MetaMap on the OHSUMED corpus, a collection of 348,566 biomedical research articles. They evaluated the method on the complete UMNSRS, MiniMayoSRS and the MayoSRS datasets; any subset information about the dataset was not explicitly stated therefore we believe a direct comparison may be possible.

In addition, a previous work very closely related to ours is a retrofitting vector method proposed by  \newcite{YuCBJW16} that incorporates ontological information into a  vector representation by including semantically related words.  In their measure, they first map a biomedical term to  MeSH terms, and second build a word vector based on the documents assigned to the respective MeSH term. They then retrofit the vector by including semantically related words found in the Unified Medical Language System. They evaluate their method on the MiniMayoSRS dataset. 

Table~\ref{tbl:relatedwork} shows a comparison to the top correlation scores reported by each of these works on the respective datasets (or subsets) they evaluated their methods on. N refers to the number of term pairs in the dataset the authors report they evaluated their method. The table also includes our top scoring results: the {\it integrated vector-res} and {\it vector-faith}. The results show that integrating semantic similarity measures into second--order co--occurrence vectors obtains a higher or on--par correlation with human judgments as the previous works reported results with the exception of the UMNSRS rel dataset. The results reported by \newcite{Pakhomov2016corpus} and \newcite{chiu2016how} obtain a higher correlation although the results can not be directly compared because both works used different subsets of the term pairs from the UMNSRS dataset.  

\section{Conclusion and Future Work}
\label{sec:conclusions}

We have presented a method for quantifying the similarity and relatedness between two terms that integrates pair--wise similarity scores into second--order vectors. The goal of this approach is two--fold. First, we  restrict the context used by the vector measure to words that exist in the biomedical domain, and second, we apply larger weights to those word pairs that are more similar to each other. Our hypothesis was that this combination would reduce the amount of noise in the vectors and therefore increase their correlation with human judgments. We evaluated our method on datasets that have been manually annotated for relatedness and similarity and found evidence to support this hypothesis. In particular we discovered that guiding the creation of a second--order context vector by selecting term pairs from biomedical text based on their semantic similarity led to improved levels of correlation with human judgment.  

We also explored using a threshold cutoff to include only those term
pairs that obtained a sufficiently large level of similarity. We found that eliminating less similar pairs improved the overall results (to a point). 
In the future, we plan to explore metrics to automatically determine the threshold cutoff appropriate for a given dataset and measure. We also plan to explore additional features that can be integrated with a second--order
vector measure that will reduce the noise but still provide
sufficient information to quantify relatedness. We are particularly
interested in approaches that learn word, phrase, and sentence
embeddings from structured corpora such as literature \cite{HillCK16}
and dictionary entries \cite{TACL711}. Such embeddings could be 
integrated into a second--order vector or be used on their own.

Finally, we compared our proposed method to other
distributional approaches, focusing on those that used
word embeddings. Our results showed that 
integrating semantic similarity measures
into second--order co--occurrence vectors obtains the
same or higher correlation with human judgments as do 
various different word embedding approaches. However, a direct
comparison was not possible due to variations in the subsets
of the UMNSRS evaluation dataset used. In the future, we
would not only like to conduct a direct comparison but
also explore integrating semantic similarity into various kinds
of word embeddings by training on pair--wise values of semantic similarity
as well as co--occurrence statistics.

\end{document}